\documentclass[review, number]{elsarticle}
\graphicspath{ {./figures/} }
\usepackage{float}
\usepackage{verbatim} %comments
\usepackage{amsmath}
\usepackage{multirow}
\usepackage{array}
\usepackage{booktabs}
\restylefloat{figure}
\floatstyle{plaintop} %table caption at top
\restylefloat{table}
\usepackage[margin=0.80in]{geometry}

\usepackage{ccaption}
\usepackage{amsmath}
\usepackage{amsbsy}
\usepackage{dcolumn}
\usepackage{multirow}
\usepackage[table]{xcolor}
\usepackage{wrapfig}
\usepackage{pifont}
\usepackage{float}
\usepackage{adjustbox}
\usepackage{wrapfig}
\usepackage{soul}
\usepackage{footnote}
\usepackage{times}
\usepackage{lipsum}
\usepackage{lineno}
\usepackage[T1]{fontenc}
\usepackage{textcomp}
\usepackage{graphicx}
\usepackage{graphics}
\usepackage{soul}

\usepackage{epstopdf}
\usepackage{caption}
\usepackage{subcaption}

\usepackage{amssymb,amstext,amsmath}
\usepackage{float}
\usepackage{booktabs}
\usepackage{array,ragged2e}
\usepackage[table]{xcolor}
\usepackage{multirow}
\usepackage{algorithm}
\usepackage{algpseudocode}
\usepackage{pifont}
\usepackage{enumerate}

\biboptions{square,numbers,sort&compress}
\journal{Digital Engineering}

\begin{document}
\begin{frontmatter}

%The title page must contain the title of the paper and the full name/full affiliation with country/e-mail address for each author and co-author of the manuscript. Please make sure you have included all elements listed below with your manuscript submission.

% \begin{titlepage}
% \begin{center}
% \vspace*{1cm}

% \textbf{ \large Expert Systems with Applications \LaTeX\ template}

% \vspace{1.5cm}

% % Author names and affiliations
% First Author$^{a,b}$ (first.author@mail.com), Second Author$^a$ (second.author@mail.com), Last Author$^c$ (last.author@mail.com) \\

% \hspace{10pt}

% % \begin{flushleft}
% % \small  
% % $^a$ Full address of first author, including the country name \\
% % $^b$ Full address of second author, including the country name \\
% % $^c$ Full address of last author, including the country name

% % \begin{comment}
% % Clearly indicate who will handle correspondence at all stages of refereeing and publication, also post-publication. Ensure that phone numbers (with country and area code) are provided in addition to the e-mail address and the complete postal address. Contact details must be kept up to date by the corresponding author.
% % \end{comment}

% % \vspace{1cm}
% % \textbf{Corresponding author at: Full address of the corresponding author, including the country name.} \\
% % Last Author \\
% % Full address of the corresponding author, including the country name \\
% % Tel: (555) 555-1234 \\
% % Email: last.author@mail.com

% % \end{flushleft}        
% \end{center}
% \end{titlepage}

\title{Analyzing Factors Influencing Driver Willingness to Accept Advanced Driver Assistance Systems}

\author[label1]{Hannah Musau \corref{cor1}}
\ead{hmusau@scsu.edu}

\author[label1]{Nana Kankam Gyimah}
\ead{ngyimah@scsu.edu}

\author[label1]{Judith Mwakalonge}
\ead{jmwakalo@scsu.edu}

\author[label2]{Gurcan Comert}
\ead{gcomert@ncat.edu}

\author[label1]{Saidi Siuhi}
\ead{ssiuhi@scsu.edu}

\cortext[cor1]{Corresponding author.}
\address[label1]{Department of Engineering, South Carolina State University, Orangeburg, South Carolina, USA, 29117}
\address[label2]{Department of Computational Engineering and Data Science, North Carolina A\&T State University, Greensboro, North Carolina, US, 27411}

\begin{abstract}
Advanced Driver Assistance Systems (ADAS) enhance highway safety by improving environmental perception and reducing human errors. However, misconceptions, trust issues, and knowledge gaps hinder widespread adoption. This study examines driver perceptions, knowledge sources, and usage patterns of ADAS in passenger vehicles. A nationwide survey collected data from a diverse sample of U.S. drivers. Machine learning models predicted ADAS adoption, with SHAP (SHapley Additive Explanations) identifying key influencing factors. Findings indicate that higher trust levels correlate with increased ADAS usage, while concerns about reliability remain a barrier. Specific features, such as Forward Collision Warning and Driver Monitoring Systems, significantly influence adoption likelihood. Demographic factors (age, gender) and driving habits (experience, frequency) also shape ADAS acceptance. Findings emphasize the influence of socioeconomic, demographic, and behavioral factors on ADAS adoption, offering guidance for automakers, policymakers, and safety advocates to improve awareness, trust, and usability.
\end{abstract}


\end{frontmatter}

\section{Introduction}
Human factors are the leading cause of road crashes, contributing to over 90\% of incidents either alone or alongside failures in vehicles or infrastructure \cite{dingus2016driver}. Risky driving behaviors arising from inadvertent errors (such as lapses in concentration and misjudgments) or deliberate traffic violations are strongly correlated with collisions \cite{young2012examining, khattak2024acceptance}. To mitigate these risks, Advanced Driver Assistance Systems (ADAS) have been developed to enhance road safety by partially automating driving functions. These technologies, including anti-lock brakes, lane-keeping assistance, and forward collision warnings, have demonstrated their potential to address human errors and improve overall road safety \cite{kwak2020association, adnan2018trust, jadaan2017connected, ahangarnejad2021review}. However, realizing these benefits hinges on overcoming critical barriers to ADAS adoption and effective utilization. While these systems have demonstrated success in controlled environments, their real-world impact depends on user acceptance, trust, and proper understanding of their functions.

The substantial safety advantages of ADAS underscore their pivotal role in minimizing collision risks and preventing injuries \cite{nandavar2023exploring}. For instance, forward collision warning systems reduce front-to-rear collisions by 27\%, low-speed autonomous emergency braking systems decrease such crashes by 43\%, and combined technologies achieve up to a 50\% reduction in collisions \cite{cicchino2017effectiveness}. Additionally, forward collision warning systems reduce near-crash incidents in fog by 35\% \cite{yue2018assessment}. Despite these benefits, real-world adoption rates remain lower than expected, raising concerns about the factors influencing user acceptance and effective utilization.

Despite its safety benefits, many drivers remain hesitant to adopt ADAS due to psychological and informational barriers, necessitating further investigation into the specific factors shaping user acceptance, awareness, and trust, as highlighted below.  
\begin{enumerate}
    \item Driver acceptance: Demographic factors, such as age, gender, and driving experience, significantly influence attitudes toward ADAS; however, existing studies present mixed and inconclusive findings regarding their impact.  
    \item Awareness and understanding: Many drivers lack sufficient knowledge of ADAS functionalities, leading to improper use, over-reliance, or disengagement from these systems.  
    \item Trust in technology: The black-box nature of ADAS machine learning models contributes to skepticism about system reliability and decision-making transparency, ultimately hindering user trust and adoption.  
\end{enumerate}

Several studies have examined the challenges associated with ADAS adoption, offering valuable insights yet leaving critical gaps. Research on \textbf{driver acceptance} has explored demographic factors such as age and gender \cite{lyu2019field}, highlighting their influence on attitudes toward ADAS; however, findings remain inconsistent, necessitating further investigation \cite{ervin2005automotive, cicchino2015experiences, eichelberger2014volvo}. Additionally, studies on \textbf{awareness and trust} indicate that inadequate understanding of ADAS functionalities often leads to risky behaviors such as over-reliance or misuse, underscoring the need for improved user education \cite{hungund2021systematic, hagl2020safe, kinosada2021trusting}. Furthermore, emerging \textbf{machine learning transparency} techniques, particularly eXplainable Artificial Intelligence (XAI) methods like SHAP, show promise in enhancing model interpretability and building trust. However, their application in ADAS adoption remains underexplored, leaving a critical gap in understanding how explainability influences user perceptions \cite{von2021transparency, ngeni2024prediction, chengula2024spatial}. \textbf{To bridge these gaps, this study leverages XAI techniques to improve ADAS transparency, enhance user comprehension, and strengthen trust, thereby facilitating safer and more effective adoption.}

This study introduces an explainable machine learning framework leveraging eXplainable Artificial Intelligence (XAI) techniques, such as SHAP, to analyze drivers’ perceptions, trust, and acceptance of Advanced Driver Assistance Systems (ADAS). The proposed framework provides actionable insights to enhance user education, increase trust, and promote safer, more informed adoption of these technologies. The key contributions of this study are as follows:

\begin{enumerate}
    \item \textbf{Driver acceptance}: By analyzing demographic and behavioral data, the framework identifies factors influencing attitudes toward ADAS, offering insights to improve acceptance.  
    \item \textbf{Awareness and understanding}: The model integrates interpretable outputs to enhance user comprehension of ADAS functionalities and benefits, reducing misuse and over-reliance.  
    \item \textbf{Trust in technology}: By employing SHAP, the framework improves transparency and reliability, addressing concerns about the opaque nature of machine learning models.  
\end{enumerate}

The remainder of this paper is structured as follows: Section $\text{II}$ reviews recent studies on advanced driver assistance systems (ADAS). Section $\text{III}$ details the data-driven methodology for ADAS acceptance, integrating predictive modeling, interpretability techniques, and text analysis. Section $\text{IV}$ presents the experimental results and discussion. Finally, Section $\text{V}$ presents the conclusion and future research directions.

\section{Related Works}
Advanced Driver Assistance Systems (ADAS) improve road safety and driving efficiency, but their adoption is influenced by multiple factors, including user acceptance, awareness, and trust in the technology. This section reviews existing research on ADAS adoption, identifying key gaps related to driver acceptance, education and awareness, and trust in ADAS transparency.

% \subsection{Driver Acceptance and Demographic Influences}
Research shows that demographic factors significantly affect ADAS adoption. Studies by \cite{davidse2006older} and \cite{wood2024exploring} indicate that older drivers generally perceive ADAS favorably due to its ability to extend independent mobility. Features such as lane-keeping assistance and forward collision warnings are particularly beneficial when adapted for sensory impairments. However, findings suggest that while older drivers recognize ADAS benefits, concerns about system complexity and interface usability hinder long-term adoption. Additionally, acceptance levels among younger drivers remain underexplored, with studies producing mixed findings on the influence of age, gender, and driving experience. Further research is needed to clarify how different demographic factors shape ADAS perceptions and long-term usage patterns.

% \subsection{Awareness, Learning Methods, and User Misuse}
Understanding how drivers learn and interact with ADAS is crucial for effective adoption. A study in Australia \cite{nandavar2023exploring} found that most drivers acquire ADAS-equipped vehicles primarily for safety but receive minimal structured training on their functionalities. Many rely on trial-and-error learning or inconsistent point-of-sale education from salespersons, leading to potential misuse, over-reliance, or disengagement. This aligns with broader concerns that inadequate awareness of ADAS capabilities contributes to risky behaviors, such as failing to override automation when necessary or misinterpreting system limitations. While some studies propose improved Human-Machine Interfaces (HMI) to enhance learning \cite{vaezipour2015reviewing}, research on structured ADAS education remains limited, highlighting a need for more comprehensive user training strategies.

% \subsection{Trust, Explainable AI, and System Transparency}
Trust in ADAS is a critical determinant of adoption, yet skepticism about system reliability persists due to the opaque, black-box nature of machine learning models. While research has examined ADAS acceptance factors, few studies have explored how AI transparency influences trust. One notable study \cite{xu2021modeling} used a random forest algorithm to analyze ADAS acceptance, identifying speed, warning duration, and driver age as key factors. However, this study did not assess trust-related concerns arising from model interpretability. Research on XAI suggests that techniques like SHAP could improve driver trust by offering more transparent explanations of system decisions, yet their application to ADAS remains underexplored. Bridging this gap is crucial for enhancing driver confidence and addressing concerns about reliability and decision-making transparency.

% \subsection{Research Gap and Justification}
Despite extensive research on ADAS adoption, significant gaps remain. Many studies focus on isolated ADAS features, such as lane departure warnings or forward collision systems, without considering the interplay between multiple functionalities. Additionally, existing research is often limited to specific driver populations, restricting the generalizability of findings. More critically, the role of XAI in ADAS acceptance remains underexplored, leaving a crucial gap in understanding the relationship between explainability and user perception.

\section{Methodology}
This study employs a data-driven approach to examine ADAS acceptance through predictive modeling, interpretability methods, and text analysis. The methodology includes structured data collection, categorical encoding, AutoML-based modeling, SHAP interpretability, and topic modeling for qualitative insights.

\subsection{Data Collection}
\label{subsection:Dataset-collection}
The study collected responses from 1,000 drivers across the United States to understand factors influencing ADAS adoption. Participants were recruited through the Qualtrics platform \cite{qualtricsQualtricsExperience} and were required to be at least 18 years old, hold a valid U.S. driver’s license, and own or have regular access to a vehicle. The survey categorized participants based on age, gender, driving experience, and familiarity with ADAS to facilitate a comprehensive analysis of adoption patterns. Table~\ref{tab:demographics} summarizes the demographic and driving characteristics used in the predictive analysis.

\begin{table}[htbp]
    \centering
    \scalebox{0.8}{
    \caption{Demographics and driving characteristics of survey respondents, including gender, age, income, vehicle type, and ADAS trust.}        \label{tab:demographics}
    \begin{tabular}{|l|l|c||l|l|c|}
        \hline
        \textbf{Variable} & \textbf{Category} & \textbf{Frequency} & \textbf{Variable} & \textbf{Category} & \textbf{Frequency} \\
        \hline
        \multirow{3}{*}{\textbf{Gender}} & Female & 601 & \multirow{5}{*}{\textbf{Driving Frequency}} & Daily & 346 \\
        & Male & 398 &  & Multiple times a day & 326 \\
        & Prefer not to say & 1 &  & Few times a week & 293 \\
        \cline{1-3}
        \multirow{6}{*}{\textbf{Age}} & 65+ & 263 &  & Rarely & 32 \\
        & 35-44 & 187 &  & Never & 3 \\
        & 55-64 & 185 & \multirow{5}{*}{\textbf{Driving Experience}} & More than 20 years & 716 \\
        & 45-54 & 178 &  & 11-20 years & 134 \\
        & 25-34 & 154 &  & 6-10 years & 84 \\
        & 18-24 & 33 &  & 1-5 years & 54 \\
        \cline{1-3}
        \multirow{6}{*}{\textbf{Race}} & White & 749 &  & Less than 1 year & 12 \\
        & Black or African American & 132 & \multirow{5}{*}{\textbf{Vehicle Type}} & Sedan & 431 \\
        & Hispanic or Latino & 60 &  & SUV & 415 \\
        & Asian & 30 &  & Truck & 85 \\
        & Multiracial & 19 &  & Other & 39 \\
        & Native American & 6 &  & Van & 30 \\
        \hline
        \multirow{3}{*}{\textbf{Education}} & High school & 385 & \multirow{8}{*}{\textbf{Employment}} & Employed Full-time & 390 \\
        & Bachelor’s degree & 372 &  & Retired & 289 \\
        & Master’s degree & 123 &  & Unemployed & 103 \\
        &  &  &  & Employed Part-time & 102 \\
        &  &  &  & Self-employed & 64 \\
        &  &  &  & Other & 43 \\
        &  &  &  & Student & 9 \\
        \hline
        \multirow{5}{*}{\textbf{Income}} & $25,000-$50,000 & 286 & \multirow{3}{*}{\textbf{Vehicle Fuel Type}} & Conventional & 898 \\
        & $\$$50,001-$\$$75,000 & 221 &  & Hybrid & 78 \\
        & Less than $\$$25,000 & 173 &  & Electric & 24 \\
        & More than $\$$100,000 & 166 & \multirow{4}{*}{\textbf{Road Type}} & Local Roads & 739 \\
        & $\$$75,001-$\$$100,000 & 154 &  & Freeways & 153 \\
        &  &  &  & Interstate & 99 \\
        &  &  &  & Other & 9 \\
        \hline
        \multirow{5}{*}{\textbf{Household Size}} & 1 & 266 & \multirow{5}{*}{\textbf{ADAS Trust Level}} & Strongly agree & 202 \\
        & 2 & 359 &  & Somewhat agree & 443 \\
        & 3 & 184 &  & Neither agree nor disagree & 249 \\
        & 4 & 114 &  & Somewhat disagree & 68 \\
        & 5 or more & 77 &  & Strongly disagree & 38 \\
        \hline
        \multirow{3}{*}{\textbf{Locale}} & Suburban & 479 &  & Urban & 284 \\
        & Rural & 237 &  &  &  \\
        \hline
    \end{tabular}
    }
\end{table}

\subsection{Data Processing}
The dataset consists of both categorical and continuous data, which were processed to create a structured dataset suitable for machine learning. Categorical variables, including \textit{Gender}, \textit{Age}, \textit{Education}, and \textit{Drive\_freq}, were analyzed and encoded appropriately:

\begin{itemize}
    \item \textbf{Ordinal Encoding}: Variables with a natural order, such as \textit{Age} and \textit{Drive\_experience}, were mapped to numerical values.
    \item \textbf{One-Hot Encoding}: Nominal variables, such as \textit{Gender} and \textit{Locale}, were transformed into binary features to prevent ordinal misinterpretation.
\end{itemize}

The \textbf{target variable}, \textit{ADAS\_Clearly}, indicated whether respondents clearly understood ADAS functionalities, with binary encoding (1 = ``Yes,'' 0 = ``No'').

\subsection{Predictive Modeling and Explainable AI (XAI) for ADAS Adoption}
This study employs an Explainable Artificial Intelligence (XAI) framework to predict and interpret ADAS adoption using AutoML-based predictive modeling and SHAP-based model interpretability.

\subsubsection{AutoGluon for Automated Machine Learning (AutoML)}
AutoGluon is a robust \textbf{AutoML framework} that automates \textit{model selection, hyperparameter tuning, and ensembling}, making it ideal for complex predictive tasks such as \textbf{Advanced Driver Assistance System (ADAS) adoption modeling}. It evaluates multiple models, optimizes hyperparameters using \textit{Bayesian Optimization and Random Search}, and employs \textit{multi-layer stacking and repeated k-fold bagging} to enhance generalization and reduce overfitting. By seamlessly handling \textbf{numerical and categorical features}, AutoGluon ensures high predictive accuracy with minimal manual intervention, making it a scalable and efficient solution for structured data modeling.

\textbf{Mathematical Formulation of AutoGluon:}

\textbf{Model Selection and Training:}  
AutoGluon automatically selects a diverse set of models and optimizes them through an ensemble approach. Given a dataset $D = \{(x_i, y_i)\}_{i=1}^{n}$ where each base model \( f_j(x) \) learns a mapping from input features to outputs as defined in Eq. (\ref{eqn:mapping}):

\begin{equation}
    \hat{y}_i^{(j)} = f_j(x_i; \theta_j)
    \label{eqn:mapping}
\end{equation}

where \( f_j(x) \) is the \( j \)-th model, \( \theta_j \) represents its trainable parameters, and \( \hat{y}_i^{(j)} \) is the predicted output for sample \( i \).

Each model is evaluated using a performance metric \( L(y, \hat{y}) \) such as cross-entropy loss (classification) or mean squared error (MSE) (regression) as defined in Eq. (\ref{eqn:loss function}):
\begin{equation}
    L(y, \hat{y}) = \frac{1}{n} \sum_{i=1}^{n} \ell(y_i, \hat{y}_i)
    \label{eqn:loss function}
\end{equation}

where \( \ell(\cdot) \) is the loss function that measures prediction error.

After selecting a diverse set of models, AutoGluon optimizes their parameters to enhance predictive performance before ensembling.

\textbf{Hyperparameter Optimization:}  
AutoGluon performs automated hyperparameter tuning by searching over a predefined hyperparameter space \( \Theta \). The optimal parameters are selected as defined in Eq. (\ref{eqn:optimal parameters}).

\begin{equation}
    \theta^* = \arg\min_{\theta \in \Theta} \mathbb{E} [L(y, f(x; \theta))]
    \label{eqn:optimal parameters}
\end{equation}

where \( \theta^* \) represents the optimal hyperparameter configuration that minimizes expected loss.

Once models are optimized, AutoGluon employs a multi-layer stacking strategy to improve predictive accuracy by leveraging diverse models.

\textbf{Multi-Layer Stacking and Ensembling:}  
AutoGluon enhances predictive performance by ensembling multiple models in a stacked learning framework. Unlike traditional ensembling methods that simply average model outputs, AutoGluon employs a \textbf{hierarchical multi-layer stacking strategy}.

Let \( f_j^{(l)}(x) \) represent a model in layer \( l \), and let \( h^{(l)}(x) \) be the stacked feature representation after \( l \) layers. The transformation follows Eq. (\ref{eqn:transformation}) as defined.

\begin{equation}
    h^{(l)}(x) = \{ f_1^{(l-1)}(x), f_2^{(l-1)}(x), ..., f_m^{(l-1)}(x) \}
    \label{eqn:transformation}
\end{equation}

where \( h^{(l)}(x) \) is the concatenation of predictions from all models in layer \( l-1 \), and each model in layer \( l \) receives both the original features and predictions from the previous layer as input.

The final prediction from the stacked ensemble is computed as defined in Eq. (\ref{eqn:stacked-ensemble}).
\begin{equation}
    \hat{y}_i = g\left( \sum_{j=1}^{m} w_j f_j(x_i) \right)
    \label{eqn:stacked-ensemble}
\end{equation}

where \( g(\cdot) \) is the meta-model that combines multiple predictions, and \( w_j \) are the learned weights of each model.

In addition to stacking, AutoGluon employs bagging techniques to further improve robustness and prevent overfitting.

\textbf{Repeated k-Fold Bagging for Robust Predictions:}  
To enhance stability and prevent overfitting, AutoGluon applies repeated k-fold bagging, ensuring that each training example is used in multiple validation sets across different model instances.

For a given dataset \( D \), AutoGluon partitions it into \( k \) folds:
\begin{itemize}
    \item Each model is trained on \( k-1 \) folds and validated on the remaining fold,
    \item This process repeats across multiple iterations to generate out-of-fold (OOF) predictions,
    \item The final ensemble aggregates these OOF predictions to improve generalization.
\end{itemize}

Mathematically, the OOF predictions for a model \( f_j \) are computed as defined in Eq. (\ref{eqn:OOF predictions}).

\begin{equation}
    \hat{y}_{i}^{\text{OOF}} = \frac{1}{k} \sum_{j=1}^{k} f_j^{(k)}(x_i)
    \label{eqn:OOF predictions}
\end{equation}

where \( f_j^{(k)} \) represents a model trained on a subset of the data.

With OOF predictions generated from bagging, AutoGluon aggregates the results to make a final prediction using weighted averaging or majority voting.

\textbf{Final Prediction Strategy:}  
Once AutoGluon completes model selection, stacking, and bagging, the final prediction \( \hat{y}_i \) is determined using either weighted averaging (for regression) or a majority vote strategy (for classification) as defined in Eq. (\ref{eqn:majority-vote-strategy}):

\begin{equation}
    \hat{y}_i = \sum_{j=1}^{m} \alpha_j f_j(x_i)
    \label{eqn:majority-vote-strategy}
\end{equation}

where \( \alpha_j \) are the optimized weights assigned to each model.

For classification problems, the final class prediction is as defined in Eq. (\ref{eqn:final class}).

\begin{equation}
    \hat{y}_i = \arg\max_c P(y_i = c \mid x_i)
    \label{eqn:final class}
\end{equation}

where \( P(y_i = c \mid x_i) \) is the probability estimated by the ensemble for class \( c \).

AutoGluon provides a powerful and efficient AutoML solution through its unique combination of model selection, hyperparameter tuning, multi-layer stacking, and repeated k-fold bagging. By automating the learning process and effectively handling mixed data types, AutoGluon is well-suited for complex predictive tasks such as \textbf{ADAS adoption modeling}, where demographic, behavioral, and psychological factors interact in nonlinear ways.

\subsubsection{SHAP for Model Interpretability}
ML-based predictions often require post-hoc interpretations to support decision-making, as interpretable models help users to understand prediction reasoning. As ML applications become more widespread, interpretability has become as critical as precision \cite{ahmad2018interpretable, sagi2020explainable}, offering transparency and valuable insights into automated predictions. Data-driven explanations, which analyze input variations without revealing the model's internal workings, quantify how these deviations affect predictions. These explanations can be grouped into three types: adversarial-based \cite{etmann2019connection, tao2018attacks}, concept-based \cite{ghorbani2019towards, zhou2018interpretable}, and perturbation-based interpretations \cite{aydin2020blotch, fong2017interpretable}, each offering unique insights into model behavior.

In this study, we use \textbf{perturbation-based interpretations} to enhance ML model transparency by masking input features and analyzing their impact on predictions. Methods like \textbf{LIME} \cite{ribeiro2016should}, \textbf{CXplain} \cite{schwab2019cxplain}, \textbf{RISE} \cite{petsiuk2018rise}, and \textbf{SHAP} \cite{lundberg2017unified} quantify feature importance differently, with LIME approximating models locally and SHAP computing \textbf{Shapley values} from \textbf{game theory} for fair attribution. However, LIME's synthetic instances may misrepresent feature values \cite{moradi2021post}, affecting interpretability. In contrast, \textbf{SHAP offers a consistent, theoretically grounded approach}, modeling input features as cooperative game players contributing to the prediction. Variants like \textbf{DeepSHAP, Kernel SHAP, LinearSHAP, and TreeSHAP} adapt it to different ML architectures. We employ \textbf{TreeSHAP}, a linear explanatory model using Shapley values, as defined in Eq. (\ref{eq:linear-exploratory}), to interpret tree-based model predictions and assess feature contributions.

\begin{align}
h(z') = \varnothing_0 + \sum_{i=1}^{N} \varnothing_i z'_i
\label{eq:linear-exploratory}
\end{align}

where h represents the explanation model, z denotes the basic features, N is the maximum size of collation and $\varnothing$ denotes the feature attribution. From Eqs. (\ref{eq:attribution}) and (\ref{eq:expected value}), the attribution of each feature is computed.

\begin{align}
\varnothing_i = \sum_{K \subseteq M \setminus \{i\}} \frac{|K|!(N - |K| - 1)!}{N!} \left[g_x(K \cup \{i\}) - g_x(K)\right]
\label{eq:attribution}
\end{align}

\begin{align}
g_x(K) = \mathbb{E}[g(x) \mid x_K]
\label{eq:expected value}
\end{align}

where term K represents a subset of the features (input), while M denotes the set of all inputs. $\mathbb{E}[g(x)$ represents the expected value of the function on subset K.

\subsection{Topic Modeling for Open-Ended Responses}
To extract key themes from open-ended ADAS feedback, this study employs \textbf{topic modeling}, a machine learning technique for identifying hidden structures in text data. Traditional methods like Latent Dirichlet Allocation (LDA) often struggle with short, sparsely worded responses. Therefore, the Gibbs Sampling Dirichlet Multinomial Mixture (GSDMM) \cite{yin2014dirichlet} model was selected for its ability to cluster such responses effectively.  

The GSDMM model assigns a single topic to each response and was configured with the following parameters:  
\begin{itemize}
    \item \textbf{K}: Maximum number of clusters, set to 5.
    \item \textbf{$\alpha$}: Probability of selecting an empty group, set to 0.1.
    \item \textbf{$\beta$}: Controls topic-word distribution, set to 0.3 to balance topic cohesion.
\end{itemize}

\section{Experimental Setting}
This section describes the dataset, the experiment setup, the hyper-parameter tuning, and the performance evaluation metrics of the network intrusion detection model.

\subsection{Dataset}
\label{subsection:Dataset}
The dataset used in this study is as described in section \ref{subsection:Dataset-collection}. This dataset is split into $70\%$ training and $30\%$ validation set to evaluate the model's performance on unseen data.

For the topic modelling on the open-ended responses, the pre-processing phase involved stop-word removal, filtering out “N/A” responses, and tokenizing text. 

% The extracted topics and their most representative words are visualized in Figure~\ref{fig:topic_analysis}.

% \subsection{Experiment Setup}

% \subsection{Hyper-parameter Tuning}
% \textcolor{violet}{The hyperparameters of the classifiers used for the pedestrian fatalities prediction are determined through a grid search and a $K$ fold cross-validation with a $K$ value of 10. The K-fold cross validation is implemented on the $80\%$ training dataset.The hyperparameters of the various classifiers are as summarized in Table \ref{table:hyperparameter setting}.}

\subsection{Evaluation Metrics}
\label{table:evaluation metrics}
The weighted average values of Precision, Recall, and F1-score, as defined in Eqs. (\ref{eq:precision}) through (\ref{eq:F1-score}), along with accuracy as defined in Eq. (\ref{eq:accuracy}), are adopted to evaluate the performance of the model. These metrics are calculated using the true positive ($tp_{i}$), true negative ($tn_{i}$), false positive ($fp_{i}$), and false negative ($fn_{i}$) values for each class $C_{i}$, where $i = 1, \cdots, m$ and $m$ represents the total number of classes in the dataset. Here, $|Y_{i}|$ denotes the total number of samples assigned to each class.

\begin{align} 
\text{Accuracy} = \frac{TP + TN}{TP + TN + FP + FN} 
\label{eq:accuracy} 
\end{align}

\begin{align}
Weighted\ Average\ Precision =\frac{\sum_{i=1}^{m}|Y_{i}| \frac{tp_{i}}{tp_{i} + fp_{i}}}{\sum_{i}^{m}|Y_{i}|} 
\label{eq:precision}
\end{align}

\begin{align}
Weighted\ Average\ Recall = \frac{\sum_{i=1}^{m}|Y_{i}| \frac{tp_{i}}{tp_{i} + fn_{i}}}{\sum_{i}^{m}|Y_{i}|} 
\label{eq:Recall}
\end{align}

\begin{align}
Weighted\ Average\ F1-score = \frac{\sum_{i=1}^{m}|y_{i}| \frac{2tp_{i}}{2tp_{i} + fp_{i} + fn_{i}}}{\sum_{i}^{m}|y_{i}|}  
\label{eq:F1-score}
\end{align}

\section{Results and Discussion}
This section presents the predictive modeling results, feature importance analysis, and qualitative insights derived from the study.

\subsection{Predictive Modeling and Explainable AI (XAI) for ADAS Adoption}
\subsubsection{AutoGluon for Automated Machine Learning (AutoML)}
Table~\ref{tab:prediction_results} presents the top ten models ranked by accuracy. The best-performing model, \textbf{WeightedEnsemble\_L2}, achieved the highest validation accuracy (0.7817), outperforming individual base models. The ensemble model, which integrates predictions from \textbf{NeuralNetTorch\_r22\_BAG\_L1} (weight: 0.875), \textbf{KNeighborsDist\_BAG\_L1} (weight: 0.062), and \textbf{CatBoost\_r137\_BAG\_L1} (weight: 0.062), highlights the strength of stacked ensembling in combining diverse learning patterns to improve generalization and robustness.  

Among the models, \textbf{NeuralNetTorch\_r22\_BAG\_L1} achieved the highest test accuracy of 0.779. \textbf{CatBoost\_r9\_BAG\_L1} and \textbf{NeuralNetTorch\_r79\_BAG\_L1} followed closely at 0.769. The strong performance of gradient boosting models, particularly \textbf{CatBoost}, suggests their suitability for this task. However, \textbf{ExtraTrees\_r42\_BAG\_L1}, despite a test accuracy of \textbf{0.765}, exhibited a lower validation accuracy (0.732), indicating potential overfitting, where the model may have memorized patterns in the training data rather than generalizing effectively.  

Beyond predictive accuracy, \textbf{computational efficiency varied significantly, influencing model selection for practical applications}. \textbf{LightGBM\_BAG\_L1} and \textbf{ExtraTrees\_r42\_BAG\_L1} exhibited the fastest training times (0.831 $s$ and 0.876 $s$, respectively), whereas \textbf{NeuralNetFastAI\_r102\_BAG\_L1} and \textbf{NeuralNetTorch models} required over 6 $s$. The trade-off between accuracy and fit time suggests that \textbf{ensemble models offer the best balance between performance and efficiency}, making them preferable for real-world applications requiring both scalability and predictive strength.

\begin{table}[h]
    \centering
    \caption{Top ten models are ranked by overall accuracy, displaying their test and validation scores, fit times, and stack levels.}
    \label{tab:prediction_results}
    \begin{tabular}{|c|l|c|c|c|c|}
        \hline
        \textbf{Rank} & \textbf{Model Name} & \textbf{Score (Test)} & \textbf{Score (Validation)} & \textbf{Fit Time} & \textbf{Stack Level} \\
        \hline
        0 & NeuralNetTorch\_r22\_BAG\_L1 & 0.779851 & 0.778491 & 6.525522 & 1 \\
        1 & WeightedEnsemble\_L2 & 0.772388 & 0.781701 & 8.4282 & 2 \\
        2 & CatBoost\_r9\_BAG\_L1 & 0.768657 & 0.772071 & 3.411079 & 1 \\
        3 & NeuralNetTorch\_r79\_BAG\_L1 & 0.768657 & 0.76244 & 6.59004 & 1 \\
        4 & CatBoost\_r177\_BAG\_L1 & 0.764925 & 0.776886 & 1.531249 & 1 \\
        5 & CatBoost\_r137\_BAG\_L1 & 0.764925 & 0.767255 & 1.611786 & 1 \\
        6 & ExtraTrees\_r42\_BAG\_L1 & 0.764925 & 0.731942 & 0.876429 & 1 \\
        7 & LightGBM\_BAG\_L1 & 0.757463 & 0.770465 & 0.831033 & 1 \\
        8 & LightGBMLarge\_BAG\_L1 & 0.757463 & 0.760835 & 2.039935 & 1 \\
        9 & NeuralNetFastAI\_r102\_BAG\_L1 & 0.757463 & 0.76565 & 7.507099 & 1 \\
        \hline
    \end{tabular}
\end{table}

Table~\ref{tab:model_performance} presents the class-wise performance metrics of the model. The model achieved a \textbf{weighted average accuracy of 0.76}, with a \textbf{recall of 0.77} and an \textbf{F1-score of 0.71}, indicating strong overall performance. However, class-wise performance indicates a disparity, particularly in the model’s ability to distinguish between positive and negative cases. The model performs well on the \textbf{dominant class (Yes)} (\textbf{0.78 accuracy, 0.98 recall, 0.87 F1-score}) but struggles with the \textbf{minority class (No)} (\textbf{0.69 accuracy, 0.14 recall, 0.23 F1-score}), reflecting a high false-negative rate and poor sensitivity in detecting negative cases.

\begin{table}[h!]
    \centering
    \caption{Class-wise accuracy, recall, and F1-score highlighting performance differences between the majority (Yes) and minority (No) classes.}
    \label{tab:model_performance}
    \begin{tabular}{|c|c|c|c|}
        \hline
        \textbf{Class} & \textbf{Accuracy} & \textbf{Recall} & \textbf{F1-Score} \\
        \hline
        0 (No) & 0.69 & 0.14 & 0.23 \\
        1 (Yes) & 0.78 & 0.98 & 0.87 \\
        \hline
        \textbf{Weighted Avg} & 0.76 & 0.77 & 0.71 \\
        \hline
    \end{tabular}
\end{table}

\subsubsection{SHAP for Model Interpretability and Feature Importance}
\begin{figure}[h!]
    \centering
    \includegraphics[width=0.7\textwidth]{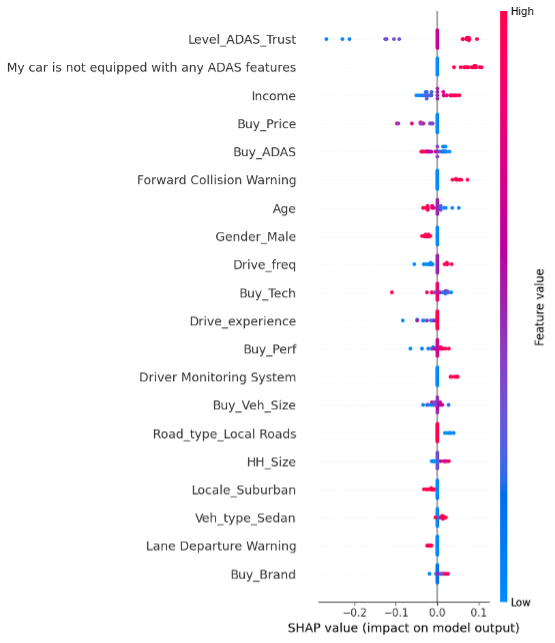} 
    \caption{The SHAP beeswarm plot illustrates feature contributions to ADAS acceptance predictions, with larger absolute SHAP values indicating greater influence. Colors represent feature values, from high (red) to low (blue).}
    \label{fig:shap_plot}
\end{figure}

To improve interpretability, we applied \textbf{SHAP (Shapley Additive Explanations)} to quantify the contribution of each feature to the model’s predictions, ensuring a fair and consistent assessment of feature importance. Figure~\ref{fig:shap_plot} presents a \textbf{SHAP beeswarm plot}, where each point represents a SHAP value for a given feature and instance. Features with larger absolute SHAP values exert greater influence on ADAS acceptance predictions.

The analysis highlights that trust in ADAS technology is the most influential factor, suggesting that \textbf{perceived reliability strongly affects adoption decisions}. The statement \textit{“My car is not equipped with any ADAS features”} also has a high impact, indicating that \textbf{prior exposure to ADAS} plays a crucial role in shaping user acceptance. Economic factors, including \textbf{income, purchase price, and willingness to invest in ADAS}, exhibit high SHAP values, emphasizing the role of \textbf{financial constraints and purchasing power} in adoption decisions.

Demographic and behavioral characteristics, such as \textbf{age, gender, driving frequency, and prior technology adoption}, contribute to model predictions but with comparatively lower influence. Additionally, features related to specific ADAS functionalities (\textit{Forward Collision Warning, Lane Departure Warning}) have a smaller impact, suggesting that trust and prior exposure outweigh individual feature preferences.

These findings underscore the critical role of consumer trust, prior exposure, and financial considerations in ADAS adoption. Understanding these factors can inform policymakers and industry stakeholders in developing strategies to enhance ADAS market penetration and user acceptance.

\subsection{Survey-Based Insights on ADAS Adoption}
To complement the predictive modeling results, this section presents survey findings on vehicle purchase considerations, awareness of ADAS features, and sources of ADAS information. These insights provide additional context on how drivers perceive and engage with ADAS technologies, helping to explain adoption patterns.

\subsubsection{Vehicle Purchase Considerations}
\begin{figure}[h!]
    \centering
    \includegraphics[width=0.85\textwidth]{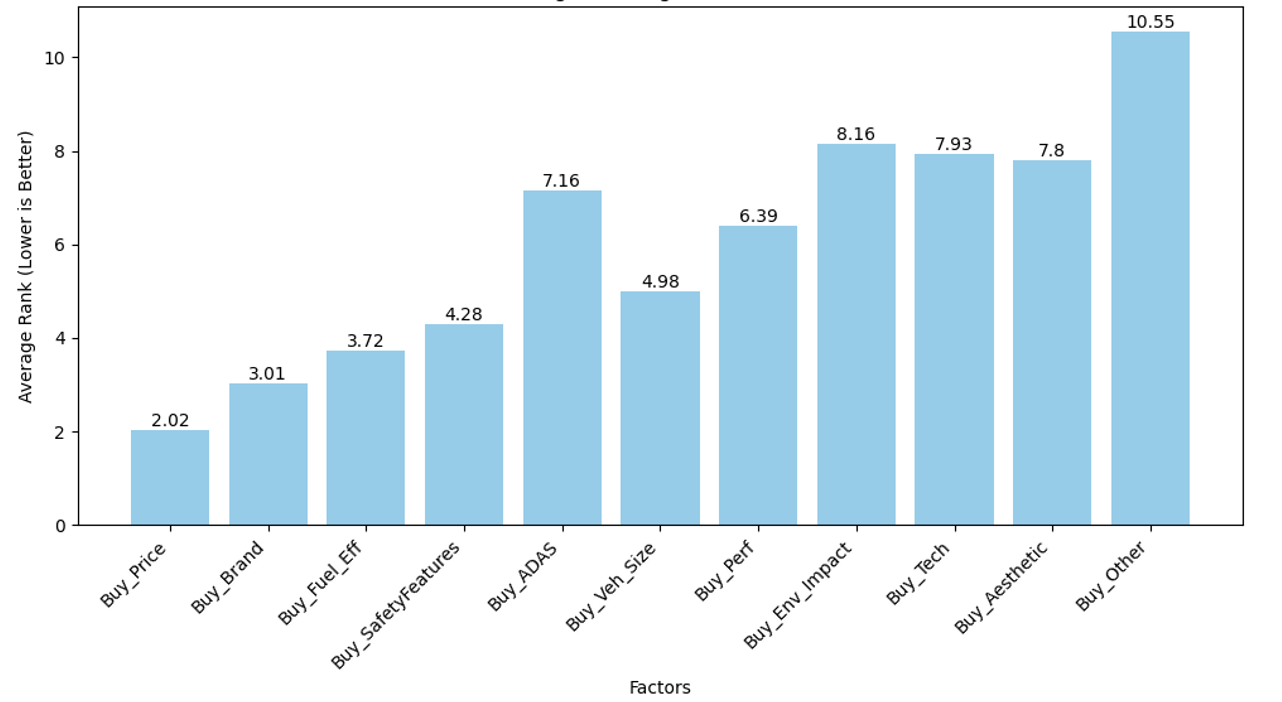} 
    \caption{Average rankings of factors considered in vehicle purchase decisions (lower ranks indicate higher importance).}
    \label{fig:rankings}
\end{figure}

Drivers ranked key factors in vehicle purchase decisions on a scale where lower values indicate greater importance, as shown in Figure~\ref{fig:rankings}. \textbf{Price} emerged as the top priority (2.02), followed by \textbf{Brand} (3.01), with \textbf{Fuel Efficiency} (3.72) and \textbf{Safety Features} (4.28) also highly valued. \textbf{ADAS features} ranked moderately in importance (7.16), positioned below \textbf{Vehicle Size} (4.98) and \textbf{Performance} (6.39) but above \textbf{Technology} (7.93) and \textbf{Aesthetic Appeal} (7.80), suggesting that while considered, they are not primary purchase drivers. \textbf{Environmental Impact} ranked lowest (8.16), indicating that while consumers prioritize affordability, brand reputation, and safety, considerations such as advanced driver assistance systems and sustainability remain secondary in purchase decisions.

These results indicate that while ADAS is a relevant factor in vehicle purchases, it is secondary to traditional concerns such as price, brand reputation, and fuel efficiency. 

\subsubsection{Awareness of ADAS Features}
\begin{figure}[h!]
    \centering
    \includegraphics[width=1\textwidth]{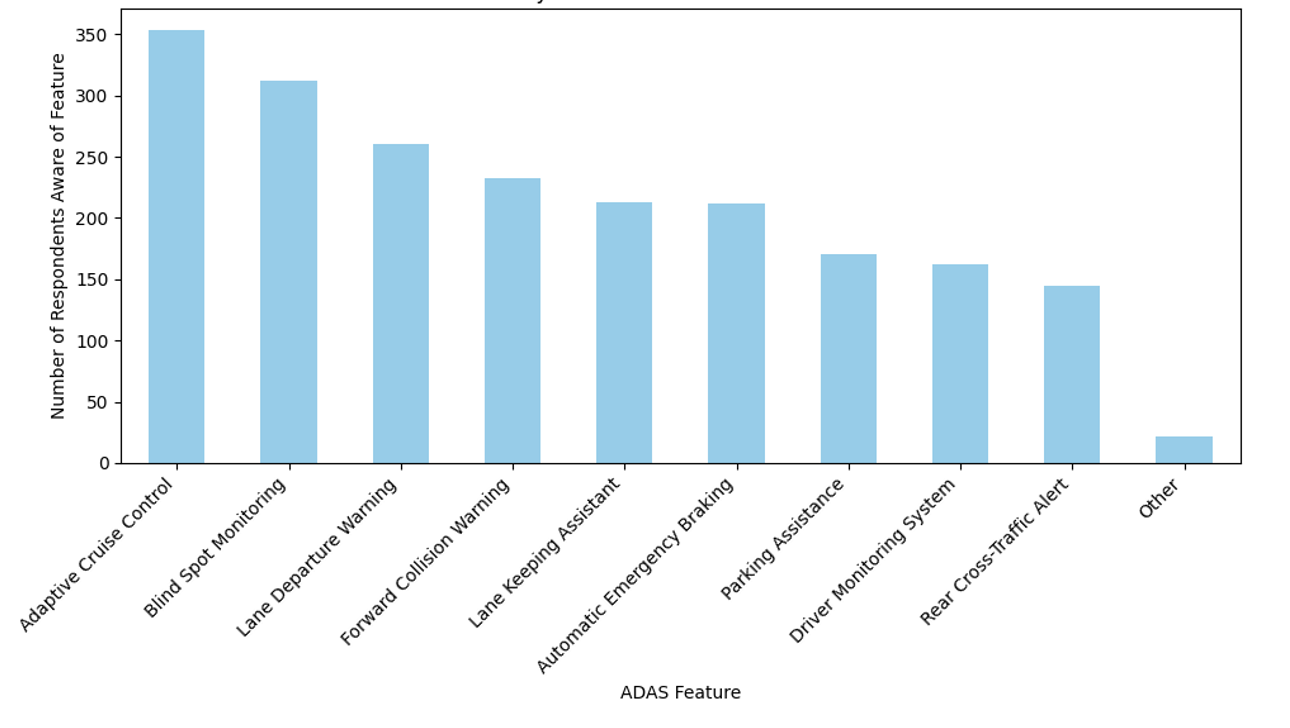} 
    \caption{Number of respondents aware of various ADAS features in vehicles equipped with ADAS.}
    \label{fig:adas_awareness}
\end{figure}

Survey respondents with ADAS-equipped vehicles identified the features available in their cars, as presented in Figure~\ref{fig:adas_awareness}. \textbf{Adaptive Cruise Control} was the most widely recognized feature, with over 350 respondents indicating awareness, followed closely by \textbf{Blind Spot Monitoring}, with just under 350 respondents. Other widely recognized features, such as \textbf{Lane Departure Warning} and \textbf{Forward Collision Warning}, were identified by approximately 300 respondents, indicating a slightly lower but still substantial level of awareness. Conversely, features such as \textbf{Lane Keeping Assist, Automatic Emergency Braking, Parking Assistance, Driver Monitoring Systems}, and \textbf{Rear Cross-Traffic Alert} had lower recognition, with significantly fewer respondents identifying these technologies, suggesting potential gaps in user familiarity.

The disparity in feature awareness suggests that some ADAS functions are more prominently marketed or intuitively understood than others. This underscores the need for better consumer education on lesser-known ADAS features to improve adoption and proper usage.

\subsubsection{Sources of ADAS Information}
\begin{figure}[h!]
    \centering
    \includegraphics[width=0.9\textwidth]{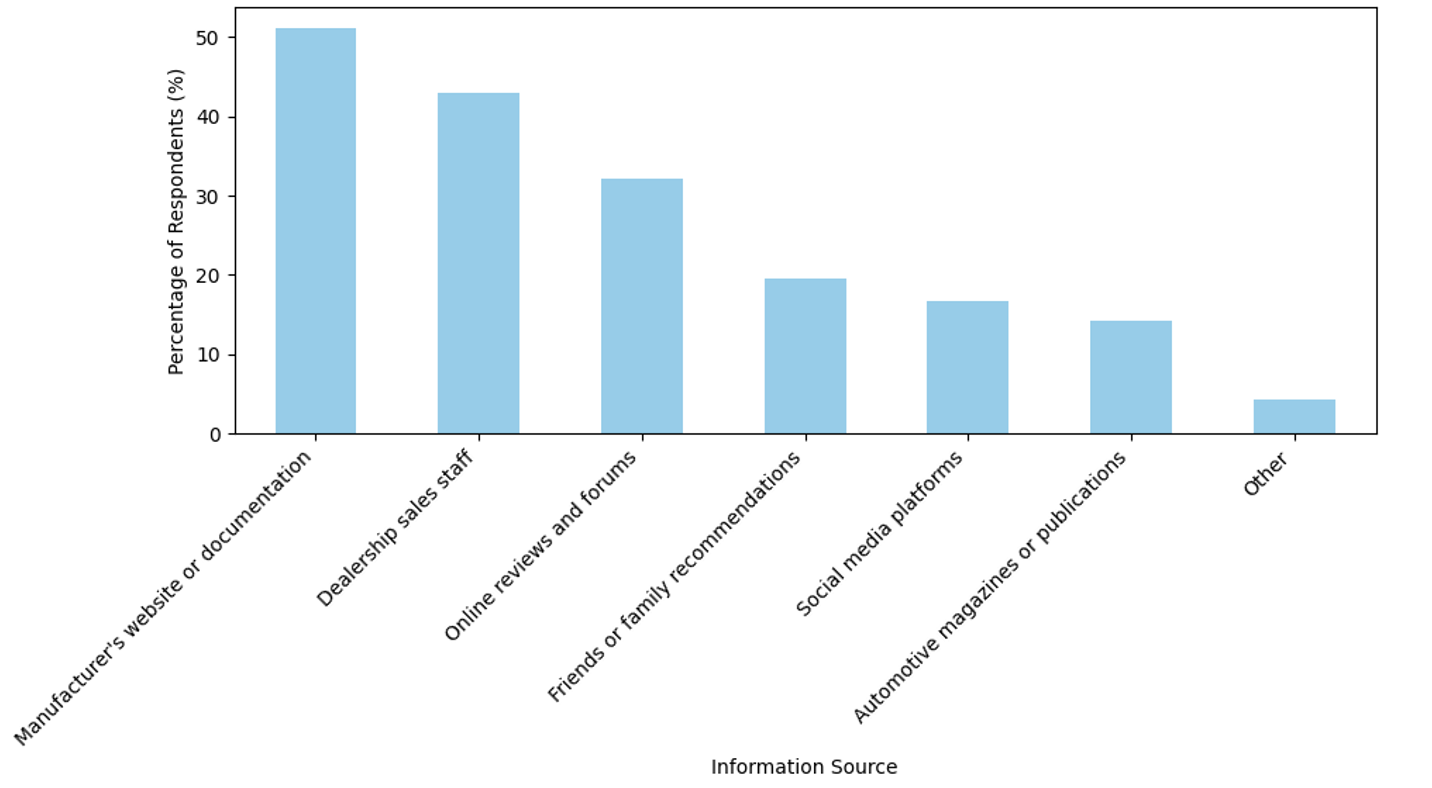}
    \caption{Primary sources of ADAS information used by respondents.}
    \label{fig:adas_info_sources}
\end{figure}

Survey participants reported their primary sources of ADAS information, as illustrated in Figure~\ref{fig:adas_info_sources}. The majority (\textbf{50\%}) obtain ADAS information from \textbf{manufacturer websites or documentation}, highlighting the central role of official sources in educating consumers. \textbf{Dealership sales staff} and \textbf{online reviews} were also significant, each cited by approximately 40\% of respondents. \textbf{Social media platforms} and \textbf{automotive magazines} were referenced by 25–30\% of respondents. \textbf{Recommendations from friends, family, and print publications} were the least utilized sources (20\%), suggesting that ADAS knowledge is primarily acquired through formal and digital channels rather than personal networks.

These findings highlight the critical role of manufacturers and dealerships in educating consumers about ADAS technologies. However, the reliance on online sources and word-of-mouth suggests gaps in structured ADAS education, which could lead to misconceptions or improper system use. Future research should investigate how to enhance information accessibility and accuracy, ensuring that consumers receive comprehensive, standardized ADAS education.

\subsection{Topic Modeling for Open-Ended Responses}
\begin{figure}[h!]
    \centering
    \includegraphics[width=0.65\textwidth]{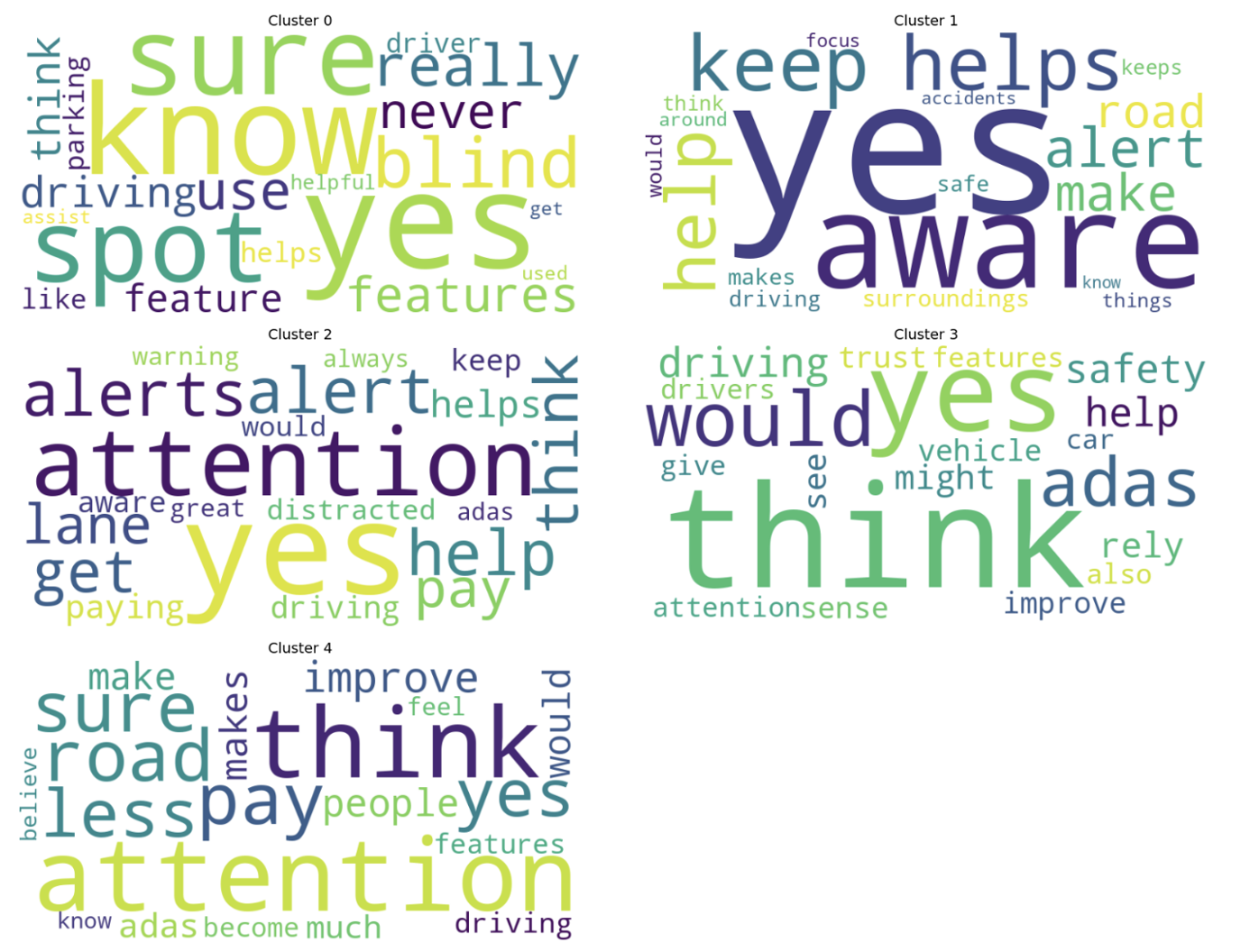} 
    \caption{Top twenty words in the top five topics extracted from open-ended responses.}
    \label{fig:topic_analysis}
\end{figure}

Participants expressed mixed views on \textbf{ADAS benefits}, with some emphasizing its role in \textbf{preventing collisions} through features like \textbf{blind-spot monitoring, lane departure alerts, and forward collision warnings}. Conversely, some respondents raised concerns about \textbf{driver over-reliance on automation}, arguing that prolonged ADAS use could \textbf{diminish situational awareness and manual driving skills}. These perceptions align with studies emphasizing the \textbf{trade-off between automation convenience and driver engagement}. As shown in Figure~\ref{fig:topic_analysis}, key themes in open-ended responses reflect \textbf{awareness, safety, attention, and trust in ADAS}. To maximize ADAS effectiveness, future educational efforts should emphasize both its \textbf{advantages and potential risks}, reinforcing the need for \textbf{continuous driver engagement and responsible system use}.

\section{Conclusion and Future Work}
This study demonstrates that drivers’ acceptance and understanding of ADAS are shaped by trust in the technology, awareness of its functionalities, and demographic characteristics. The analysis reveals that higher trust levels, prior exposure to ADAS, and purchase-related preferences—such as prioritizing vehicle technology—are strongly associated with greater acceptance. However, several challenges persist, including age-related barriers to ADAS adoption, concerns about driver over-reliance on automation, and skepticism regarding potential distractions or unintended consequences. The diverse perceptions reflected in the open-ended responses highlight the need for a balanced approach to ADAS implementation and education, ensuring that drivers remain actively engaged while benefiting from the safety enhancements these systems provide.  

Future research should develop targeted educational strategies, including interactive tools like augmented reality and virtual training, to enhance ADAS comprehension and trust. Studies should examine ADAS adoption barriers among older adults and infrequent drivers to develop tailored support mechanisms. Another key area is advancing adaptive ADAS technologies that adjust functionalities based on environmental conditions and driver behavior to improve safety and usability. Finally, longitudinal studies should assess ADAS impacts on over-reliance and disengagement to ensure these systems enhance road safety.  

\section*{Credit authorship contribution statement}
\noindent \textbf{Hannah Musau}: Conceptualization, Writing, methodology, and Data analysis. \textbf{Nana Kankam Gyimah}: Conceptualization, Writing, methodology, and Data analysis. \textbf{Judith Mwakalonge}: Conceptualization, Writing, methodology, and Data analysis. \textbf{Gurcan Comert}: Conceptualization, Writing, methodology, and Data analysis. \textbf{Saidi Siuhi}: Conceptualization, Writing, methodology, and Data analysis.

\section*{Declaration of Competing Interest}
The authors declare that there are no competing financial interests or personal relationships that could have influenced the work reported in this paper.

\section*{Acknowledgements}
This research was supported by the U.S. Department of Education through Grant No. P382G320015, administered by the Transportation Program at South Carolina State University (SCSU), the Center for Connected Multimodal Mobility (C2M2) and the National Center for Transportation Cybersecurity and Resiliency (TraCR), USA, headquartered at Clemson University, Clemson, South Carolina, Department of Energy Minority Serving Institutions Partnership Program (MSIPP) managed by the Savannah River National Laboratory under BSRA contract TOA 0000525174 CN1, MSEIP II Cyber Grants: P120A190061, P120A210048, FM-MHP-0678-22-01-00, USA, National Science Foundation (NSF), USA Grants Nos. 1954532, 2131080, 2200457, OIA-2242812, 2234920, and 2305470. Any opinions, findings, conclusions, or recommendations expressed in this material are those of the authors and do not necessarily reflect the views of the C2M2 or TraCR and the official policy or position of the USDOT/OST-R, or any State or other entity. The U.S. Government assumes no liability for the contents or use thereof.

\section*{Data availability}
The data used in this study was collected through experiments designed and conducted by the research team. Due to privacy and ethical considerations involving human participants, the data cannot be shared publicly. However, interested researchers may contact the corresponding author to request access, subject to approval and compliance with relevant data protection regulations.

\bibliography{sample}

\end{document}